\documentclass{article}

\usepackage{microtype}
\usepackage{graphicx}
\usepackage{booktabs} 
\graphicspath{{figures/}}
\usepackage{natbib}
\usepackage{enumitem}
\usepackage{hyperref}

\usepackage[accepted]{icml2018}

\usepackage{tikz}

\usepackage{amsmath}
\usepackage{amssymb,amsfonts,bm}
\usepackage{amsthm}
\usepackage{varwidth}
\usepackage{graphicx, color}
\usepackage{tikz}
\usetikzlibrary{shapes,arrows}
\usepackage{verbatim}
\usepackage{subcaption}
\usepackage[flushleft]{threeparttable}
 \newtheorem{Theorem}{Theorem}
\usetikzlibrary{arrows,positioning, decorations.pathreplacing,decorations.pathmorphing}
\tikzset{
    >=stealth',
    punkt/.style={
           rectangle,
           rounded corners,
           draw=black, very thick,
           text width=6.5em,
           minimum height=2em,
           text centered},
    pil/.style={
           ->,
           thick,
           shorten <=2pt,
           shorten >=2pt,},
    decoration={brace},
  	tuborg/.style={decorate},
	tubnode/.style={midway, right=2pt}
}
\pgfdeclarelayer{background}
\pgfdeclarelayer{foreground}
\pgfsetlayers{background,main,foreground}
\tikzstyle{block} = [draw, text width=4em, text centered, fill=blue!20, rectangle,
    minimum height=3em, minimum width=4em]
\tikzstyle{sum} = [draw, fill=blue!20, circle, node distance=0.01cm]
\tikzstyle{input} = [coordinate]
\tikzstyle{output} = [coordinate]
\tikzstyle{pinstyle} = [pin edge={to-,thin,black}]
\tikzstyle{blockAtt} = [draw, fill=red!20, rectangle,
    minimum height=3em, minimum width=4em,text width=4em, text centered]
\tikzstyle{annot} = [text width= 3em, text centered]

\icmltitlerunning{Generative Adversarial Privacy}

\begin{document}
	
\setlength{\belowdisplayskip}{0.5pt}

\setlength{\parskip}{0.2\baselineskip}

\twocolumn[
\icmltitle{Generative Adversarial Privacy}



\icmlsetsymbol{equal}{*}
\begin{icmlauthorlist}
\icmlauthor{Chong Huang}{As}
\icmlauthor{Peter Kairouz}{St}
\icmlauthor{Xiao Chen}{St}
\icmlauthor{Lalitha Sankar}{As}
\icmlauthor{Ram Rajagopal}{St}

\icmlaffiliation{As}{Arizona State University, Tempe, AZ, USA}
\icmlaffiliation{St}{Stanford University, Stanford, CA, USA}
\end{icmlauthorlist}

\icmlcorrespondingauthor{Lalitha Sankar}{lalithasankar@asu.edu}

\icmlkeywords{Data Privacy, Differential Privacy, Adversarial Learning, Generative Adversarial Networks, Minimax Games, Information Theory}

\vskip 0.3in
]



 \printAffiliationsAndNotice{}  

\begin{abstract}
We present a data-driven framework called generative adversarial privacy (GAP). Inspired by recent advancements in generative adversarial networks (GANs),  GAP allows the data holder to learn the privatization mechanism directly from the data. Under GAP, finding the optimal privacy mechanism is formulated as a constrained minimax game between a privatizer and an adversary. We show that for appropriately chosen adversarial loss functions, GAP provides privacy guarantees against strong information-theoretic adversaries. We also evaluate GAP's performance on the GENKI face database.
\end{abstract}

\section{Introduction}
The use of machine learning algorithms for data analytics has recently seen unprecedented success for a variety of problems of practical relevance such as image classification, natural language processing, and prediction of consumer behavior, electricity use, political preferences, to name a few. The success of these algorithms hinges on the availability of large datasets, which are often crowd-sourced and contain private information.
This, in turn, has led to privacy concerns and a growing body of research focused on developing privacy-guaranteed learning techniques.

Moving towards randomization-based methods, in recent years, two distinct approaches with provable \textit{statistical privacy} guarantees have emerged: (a) context-free approaches that assume worst-case dataset statistics and adversaries; (b) context-aware approaches that explicitly model the dataset statistics and adversary's capabilities. On the one hand, context-free approaches (such as differential privacy \cite{Dwork2014}) provide strong privacy guarantees against worst-case adversaries, but often lead to a significant reduction in the utility and increased sample complexity \cite{Fienberg2010,WangLeeKifer2015,YuFienbergSlavkovicUhler2014,KarwaSlavkovic2016,DuchiWainwrightJordan2016,Kairouz2016a}. On the other, context-aware approaches (such as mutual information privacy \cite{Rebollo-Monedero2010,PinCalmon2012,Sankar_TIFS_2013}) achieve a better privacy-utility tradeoff by incorporating the statistics of the dataset and explicitly modeling the public and private variables, but necessitate knowledge of data statistics (such as joint priors over the public and private variables). Such information is hardly ever present in practice.

Given the challenges of existing approaches, we take a fundamentally new approach towards enabling private data publishing. Instead of adopting worst-case, context-free notions of data privacy, we introduce a novel context-aware model of privacy that allows the designer to cleverly add noise where it matters. We overcome the issue of statistical knowledge by taking a \textit{data-driven approach}; specifically, we leverage recent advancements in generative adversarial networks (GANs) \citep{ goodfellow2014generative,mirza2014conditional} to introduce a framework for context-aware privacy that we call \textit{generative adversarial privacy} (GAP). 

%

\section{Generative Adversarial Privacy}
\label{sec:model}

We consider a dataset $\mathcal{D}$ which contains both public and private variables for $n$ individuals. We represent the public variables by a random variable $X$, and the private variables (which are typically correlated with the public variables) by a random variable $Y$. Each dataset entry contains a pair of public and private variables denoted by $(X,Y)$. We assume that each entry pair $(X,Y)$ is distributed according to $P(X,Y)$, and is independent from other entry pairs in the dataset. We define the privacy mechanism as a randomized mapping given by $\hat{X} = g(X,Y)$. 


We define $\hat{Y}=h(g(X,Y))$ to be the adversary's inference of the private variable $Y$ from $\hat{X}$ using a decision rule $h$. We allow for \textit{hard decision rules} under which $h(g(X,Y))$ is a direct estimate of $Y$ and  \textit{soft decision rules} under which $h(g(X,Y)) = P_{h}(\cdot|g(X,Y))$ is a distribution over $\mathcal{Y}$. To quantify the adversary's performance, we use a loss function $\ell(h(g(X=x)),Y=y)$ defined for every public-private pair $(x,y)$. Thus, the {expected loss} of the adversary with respect to (\textit{w.r.t.}) $X$ and $Y$ is
\begin{equation}
	\label{eq:systemadversaryloss}
	L(h, g)\triangleq \mathbb{E}[\ell(h(g(X,Y)),Y)],
\end{equation}
where the expectation is taken over $P(X,Y)$ and the randomness in $g$ and $h$.

The data holder would like to find a privacy mechanism $g$ that is both privacy preserving (in the sense that it is difficult for the adversary to learn $Y$ from $\hat{X}$) and utility preserving (in the sense that it does not distort the original data too much). In contrast, for a fixed choice of privacy mechanism $g$, the adversary would like to find a (potentially randomized) function $h$ that {minimizes its expected loss, which is equivalent to maximizing the negative of the expected loss}. This leads to a constrained minimax game between the privatizer and the adversary given by
\begin{align}
	\label{eq:generalopt}
	\min_{g(\cdot)}\max_{h(\cdot)} \quad & -L(h,g) \\ \nonumber
	s.t.  \quad & \mathbb{E}[d(g(X,Y),X)]\le D,
\end{align}
where the constant $D\ge0$ determines the allowable distortion for the privatizer and the expectation is taken over $P(X,Y)$ and the randomness in $g$ and $h$.

\begin{Theorem}
	\label{thm:GAP}
	Under the class of hard decision rules, when $\ell(h(g(x,y),y))$ is the 0-1 loss function, the GAP minimax problem in \eqref{eq:generalopt} simplifies to
	\begin{align}
		\label{eq:GAP_MAP}
		\min\limits_{g(\cdot)} \quad & \max_{y \in \mathcal{Y}} P(y, g(X,Y))  \\ \nonumber
		s.t.  \quad & \mathbb{E}[d(g(X,Y),X)]\le D,
	\end{align}
	indicating that maximizing the probability of correctly guessing $Y$ is the optimal adversarial strategy for any privatizer, i.e., the adversary uses the MAP decision rule.  On the other hand, for a soft-decision decoding adversary (i.e., $h = P_h(y|\hat{x})$ is a distribution over $\mathcal{Y}$) under log-loss function $\ell(h(g(X,Y)),y) = \log \frac{1}{P_h(y|g(X,Y))}$, the optimal adversarial strategy $h^*$ is the posterior belief of $Y$ given $g(X,Y)$  and the GAP minimax problem in \eqref{eq:generalopt} is equivalent to
	\begin{align}
		\label{eq:GAP_MI}
		\min\limits_{g(\cdot)} \quad & I(g(X,Y);Y) \\ \nonumber
		s.t.  \quad & \mathbb{E}[d(g(X,Y),X)]\le D,
	\end{align}
	where $I(g(X,Y);Y)$ is the mutual information (MI) between $g(X,Y)$ and $Y$.
\end{Theorem}


The above theorem shows that GAP can can recover MI privacy (under a log loss) and MAP privacy (under a 0-1 loss). The proof of Theorem \ref{thm:GAP} is omitted for brevity.

\subsection{Data-driven GAP}
\label{sec:learningadversary}
In the absence of $P(X,Y)$, we propose a data-driven version of GAP that allows the data holder to learn privatization mechanisms directly from a dataset $\mathcal{D} = \{(x_{(i)},y_{(i)})\}_{i = 1}^{n}$. Under the data-driven version of GAP, we represent the privacy mechanism via a generative model $g(X,Y;\theta_{p})$ parameterized by $\theta_{p}$. In the training phase, the data holder learns the optimal parameters $\theta_p$ by competing against a \textit{computational adversary}: a classifier modeled by a neural network $h(g(X,Y;\theta_{p}); \theta_a)$ parameterized by $\theta_{a}$.


In the data-driven approach, we can quantify the adversary's {\textit{empirical loss}} by
\begin{align}
	\label{eq:lossCEbinary}
	L_{n}(\theta_p,\theta_a)= & -\frac{1}{n}\sum\limits_{i=1}^{n} \ell(h(g(x_{(i)},y_{(i)}; \theta_{p});\theta_{a}), y_{(i)})
\end{align}
where $(x_{(i)}, y_{(i)})$ is the $i^{th}$ row of $\mathcal{D}$. The optimal parameters for the privatizer and adversary are the solutions to
\begin{align}
	\label{eq:learnedprivatizer}
	\min_{\theta_{p}}\max\limits_{\theta_a}\quad & -L_{{n}}(\theta_p,\theta_a) \\ \nonumber
	s.t.  \quad&  \mathbb{E}_{\mathcal{D}}[d(g(X,Y;\theta_{p}), X)]\le D,
\end{align}
where the expectation is over $\mathcal{D}$ and the randomness in $g$.

\begin{figure}
	\def\layersep{2cm}
	\centering
	\resizebox{8cm}{2.1cm}{
		\begin{tikzpicture}[shorten >=1pt,->,draw=black!50, node distance=\layersep]
		
		\tikzstyle{every pin edge}=[<-,shorten <=1pt]
		\tikzstyle{neuron}=[circle,fill=black!25,minimum size=17pt,inner sep=0pt]
		\tikzstyle{input neuron}=[neuron, fill=green!50];
		\tikzstyle{output neuron}=[neuron, fill=red!50];
		\tikzstyle{hidden neuron}=[neuron, fill=blue!50];
		\tikzstyle{annot} = [text width=6em, text centered]
		

		\node[input neuron, yshift=0.5cm, pin={[pin edge={<-}]left:$({X}, {Y})$}] (I-1) at (-3,-1) {};
		
		\node[input neuron,yshift=0.5cm,  pin={[pin edge={<-}]left:Noise sequence}] (I-2) at (-3,-2) {};
		
		\foreach \name / \y in {1,...,4}
		\path[yshift=0.5cm]
		node[hidden neuron] (H1-\name) at (\layersep -3cm,28-\y cm) {};

		\foreach \name / \y in {1,...,4}
		\path[yshift=0.5cm]
		node[hidden neuron] (H-\name) at (\layersep,28-\y cm) {};
		
		\node[output neuron, right of=H-2, yshift=-0.5cm] (O) {};
		
		\foreach \source in {1,...,2}
		\foreach \dest in {1,...,4}
		\path (I-\source) edge (H1-\dest);

		\foreach \source in {1,...,4}
		\foreach \dest in {1,...,4}
		\path (H1-\source) edge (H-\dest);
		\foreach \source in {1,...,4}
		\path (H-\source) edge (O);

		\node[input neuron, right of=O] (AI) {};
		
		\foreach \name / \y in {1,...,4}
		\node[hidden neuron, right of=AI, yshift=1.5cm] (A-\name) at (6,-\y) {};
		
		\foreach \name / \y in {1,...,2}
		\path[yshift=0.5cm]
		node[hidden neuron,  right of =A, xshift=-0.5cm] (B-\name) at (8,-\y cm) {};
		
		\node[output neuron,pin={[pin edge={->}]right:$\hat{Y}$}, right of=AI, xshift=2.5cm] (OA) {};
		
		\path (O) edge (AI);
		\foreach \dest in {1,...,4}
		\path (AI) edge (A-\dest);
		
		\foreach \source in {1,...,4}
		\foreach \dest in {1,...,2}
		\path (A-\source) edge (B-\dest);
		
		\foreach \source in {1,...,2}
		\path (B-\source) edge (OA);
		\node[annot, right of=H-4](NI){};
		\node[annot,above of=H-1, node distance=1cm, xshift=-1.7cm] (pn) {Privatizer};
		\node[annot,above of=A-1, node distance=1cm] (hl) {Adversary};
		
		\node[annot,right of=O, node distance=1cm, yshift=0.2cm] (xhat) {$\hat{{
					X}}$};
		\end{tikzpicture}
	}
	\caption{A multi-layer neural network model for the privatizer and adversary}
	\label{fig:ANN}
\end{figure}
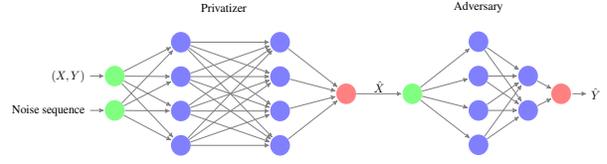

\section{GAP for the GENKI Dataset}
\label{sec:genki}
To demonstrate GAP's capability of learning the privacy mechanism directly from the data, we test our model on the GENKI dataset \cite{whitehill2012discriminately} which contains 1940 grey-scale images of faces. We consider gender as private variable $Y$ and the image pixels as public variable $X$. Two different privatizer architectures are studied: the feedforward neural network privatizer (FNNP) and the transposed convolutional neural network privatizer (TCNNP). The FNNP uses a five-layer feedforward neural network to combine the low-dimensional noise with the original image. The TCNNP uses a three-layer transposed convolutional neural network to generate high-dimensional additive noise from low-dimensional noise. The adversary is modeled by a seven-layer convolutional neural network.

\begin{figure}
	\centering
	\includegraphics[width=0.89\columnwidth]{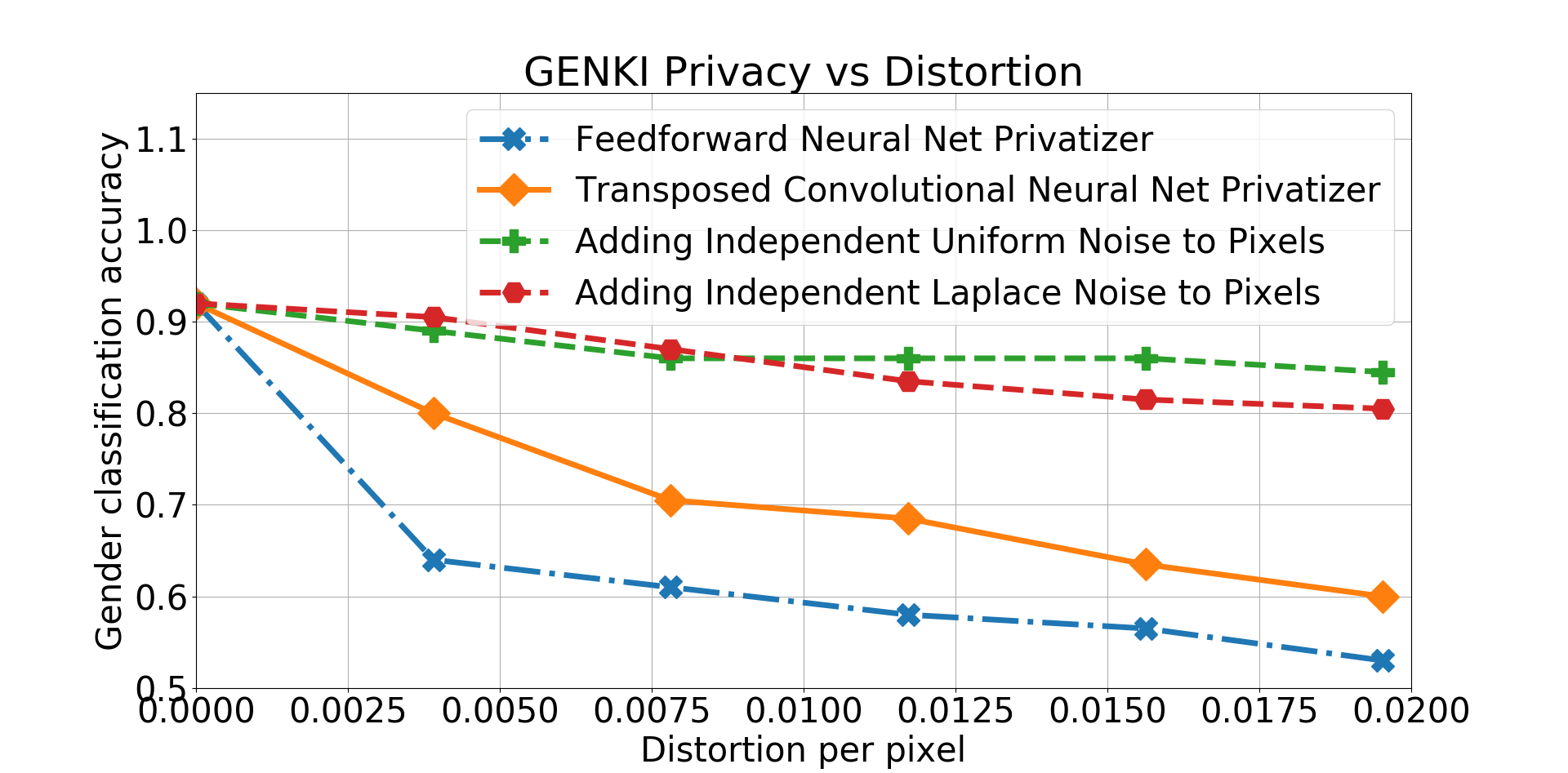}
	\caption{Privacy-distortion tradeoff}
	\label{fig:genkiprivacy}
\end{figure}

\begin{figure}
	\centering
	\includegraphics[width=0.69\columnwidth]{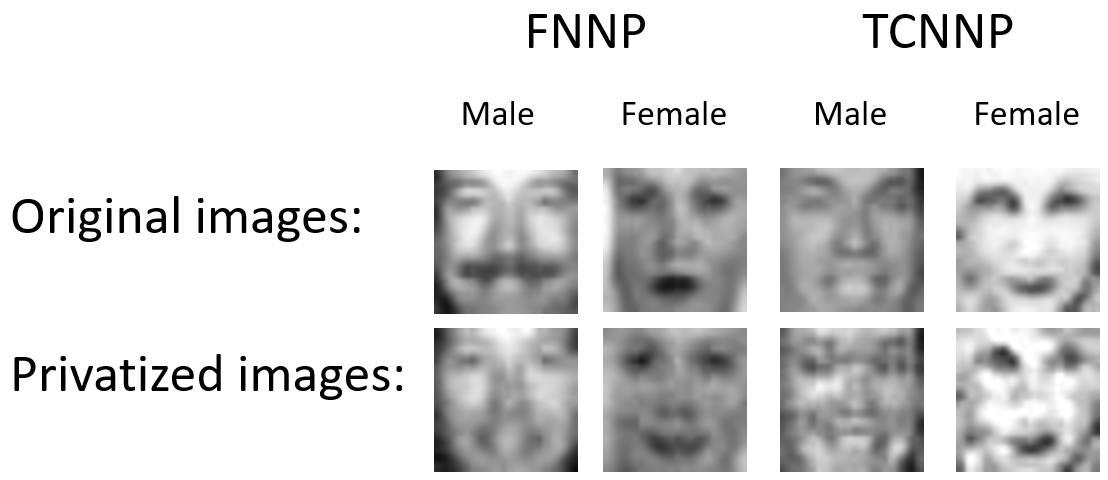}
	\caption{Miss-classified privatized image samples}
	\label{fig:privatizedimages}
\end{figure}

Figure~\ref{fig:genkiprivacy} illustrates the gender classification accuracy of the adversary for different values of distortion. We observe that the adversary's accuracy of classifying the private label (gender) decreases progressively as the distortion increases. Given the same distortion value, FNNP achieves better privacy protection compared with TCNNP. 
The adversary's miss-classified privatized image samples are shown in Figure \ref{fig:privatizedimages}. We observe that both privatizers change mostly eyes, nose, mouth, beard, and hair. 




\bibliographystyle{icml2018}
\bibliography{sample,references,LS_Privacy_Refs,StatLearning_Bibliography,Hands_Dataset_ref}

\begin{thebibliography}{13}
\providecommand{\natexlab}[1]{#1}
\providecommand{\url}[1]{\texttt{#1}}
\expandafter\ifx\csname urlstyle\endcsname\relax
  \providecommand{\doi}[1]{doi: #1}\else
  \providecommand{\doi}{doi: \begingroup \urlstyle{rm}\Url}\fi

\bibitem[Calmon \& Fawaz(2012)Calmon and Fawaz]{PinCalmon2012}
Calmon, F.~P. and Fawaz, N.
\newblock Privacy against statistical inference.
\newblock In \emph{Communication, Control, and Computing (Allerton), 2012 50th
  Annual Allerton Conference on}, pp.\  1401--1408, 2012.

\bibitem[Duchi et~al.(2016)Duchi, Wainwright, and
  Jordan]{DuchiWainwrightJordan2016}
Duchi, John, Wainwright, Martin, and Jordan, Michael.
\newblock Minimax optimal procedures for locally private estimation.
\newblock \emph{arXiv preprint arXiv:1604.02390}, 2016.

\bibitem[Dwork \& Roth(2014)Dwork and Roth]{Dwork2014}
Dwork, Cynthia and Roth, Aaron.
\newblock The algorithmic foundations of differential privacy.
\newblock \emph{Found. Trends Theor. Comput. Sci.}, 9\penalty0 (3--4):\penalty0
  211--407, August 2014.
\newblock ISSN 1551-305X.
\newblock \doi{10.1561/0400000042}.
\newblock URL \url{http://dx.doi.org/10.1561/0400000042}.

\bibitem[Fienberg et~al.(2010)Fienberg, Rinaldo, and Yang]{Fienberg2010}
Fienberg, Stephen~E., Rinaldo, Alessandro, and Yang, Xiaolin.
\newblock \emph{Differential Privacy and the Risk-Utility Tradeoff for
  Multi-dimensional Contingency Tables}, pp.\  187--199.
\newblock Springer Berlin Heidelberg, Berlin, Heidelberg, 2010.
\newblock ISBN 978-3-642-15838-4.
\newblock \doi{10.1007/978-3-642-15838-4_17}.
\newblock URL \url{https://doi.org/10.1007/978-3-642-15838-4_17}.

\bibitem[Goodfellow et~al.(2014)Goodfellow, Pouget-Abadie, Mirza, Xu,
  Warde-Farley, Ozair, Courville, and Bengio]{goodfellow2014generative}
Goodfellow, Ian, Pouget-Abadie, Jean, Mirza, Mehdi, Xu, Bing, Warde-Farley,
  David, Ozair, Sherjil, Courville, Aaron, and Bengio, Yoshua.
\newblock Generative adversarial nets.
\newblock In \emph{Advances in neural information processing systems}, pp.\
  2672--2680, 2014.

\bibitem[Kairouz et~al.(2016)Kairouz, Bonawitz, and Ramage]{Kairouz2016a}
Kairouz, Peter, Bonawitz, Keith, and Ramage, Daniel.
\newblock Discrete distribution estimation under local privacy.
\newblock In \emph{Proceedings of the 33rd International Conference on
  International Conference on Machine Learning - Volume 48}, ICML'16, pp.\
  2436--2444. JMLR.org, 2016.
\newblock URL \url{http://dl.acm.org/citation.cfm?id=3045390.3045647}.

\bibitem[Karwa \& Slavkovi{\'c}(2016)Karwa and
  Slavkovi{\'c}]{KarwaSlavkovic2016}
Karwa, Vishesh and Slavkovi{\'c}, Aleksandra.
\newblock Inference using noisy degrees: Differentially private $\beta$-model
  and synthetic graphs.
\newblock \emph{The Annals of Statistics}, 44\penalty0 (1):\penalty0 87--112,
  2016.

\bibitem[Mirza \& Osindero(2014)Mirza and Osindero]{mirza2014conditional}
Mirza, Mehdi and Osindero, Simon.
\newblock Conditional generative adversarial nets.
\newblock \emph{arXiv preprint arXiv:1411.1784}, 2014.

\bibitem[Rebollo-Monedero et~al.(2010)Rebollo-Monedero, Forne, and
  Domingo-Ferrer]{Rebollo-Monedero2010}
Rebollo-Monedero, D., Forne, J., and Domingo-Ferrer, J.
\newblock From t-{Closeness}-{Like} {Privacy} to {Postrandomization} via
  {Information} {Theory}.
\newblock \emph{IEEE Transactions on Knowledge and Data Engineering},
  22\penalty0 (11):\penalty0 1623--1636, November 2010.
\newblock ISSN 1041-4347.
\newblock \doi{10.1109/TKDE.2009.190}.

\bibitem[Sankar et~al.(2013)Sankar, Rajagopalan, and Poor]{Sankar_TIFS_2013}
Sankar, L., Rajagopalan, S.~R., and Poor, H.~V.
\newblock Utility-privacy tradeoffs in databases: An information-theoretic
  approach.
\newblock \emph{IEEE Transactions on Information Forensics and Security},
  8\penalty0 (6):\penalty0 838--852, 2013.

\bibitem[Wang et~al.(2015)Wang, Lee, and Kifer]{WangLeeKifer2015}
Wang, Yue, Lee, Jaewoo, and Kifer, Daniel.
\newblock Differentially private hypothesis testing, revisited.
\newblock \emph{arXiv preprint arXiv:1511.03376}, 2015.

\bibitem[Whitehill \& Movellan(2012)Whitehill and
  Movellan]{whitehill2012discriminately}
Whitehill, Jacob and Movellan, Javier.
\newblock Discriminately decreasing discriminability with learned image
  filters.
\newblock In \emph{Computer Vision and Pattern Recognition (CVPR), 2012 IEEE
  Conference on}, pp.\  2488--2495. IEEE, 2012.

\bibitem[Yu et~al.(2014)Yu, Fienberg, Slavkovi{\'c}, and
  Uhler]{YuFienbergSlavkovicUhler2014}
Yu, Fei, Fienberg, Stephen~E, Slavkovi{\'c}, Aleksandra~B, and Uhler, Caroline.
\newblock Scalable privacy-preserving data sharing methodology for genome-wide
  association studies.
\newblock \emph{Journal of biomedical informatics}, 50:\penalty0 133--141,
  2014.

\end{thebibliography}

%
%

\end{document}